# Vehicle Tracking Through Magnetic Sensors: A Case Study of Two-lane Road

Xiaojiang Ren[a,b], Yan Wang[a], and YingFan Geng[a]

[a] Xidian University, Xi'an, China
[b] Xidian University Guangzhou Institute of Technology, GuangZhou, China

## ABSTRACT

Intelligent Transportation Systems (ITS) have a pressing need for efficient and reliable traffic surveillance solutions. This paper for the first time proposes a surveillance system that utilizes low-cost magnetic sensors for detecting and tracking vehicles continuously along the road. The system uses multiple sensors mounted along the roadside and lane boundaries to capture the movement of vehicles. Real-time measurement data is collected by base stations and processed to produce vehicle trajectories that include position, timestamp, and speed. To address the challenge of tracking vehicles continuously on a road network using a large amount of unlabeled magnetic sensor measurements, we first define a vehicle trajectory tracking problem. We then propose a graph-based data association algorithm to track each detected vehicle, and design a related online algorithm framework respectively. We finally validate the performance via both experimental simulation and real-world road deployment. The experimental results demonstrate that the proposed solution provides a cost-effective solution to capture the driving status of vehicles and on that basis form various traffic safety and efficiency applications.

## 1. INTRODUCTION

Intelligent Transportation Systems(ITS) integrate advanced electronic technology, information technology and system engineering technology into the traffic surveillance and management using various road facilities.[5]

Magnetic sensor has recently been utilized for traffic surveillance due to its advantages of easy installation, low cost, small size, strong anti-jamming capability etc.[2,18,19] Researchers worldwide have studied the application of magnetic sensor for traffic surveillance and smart transportation, including vehicle detection and classification, vehicle counting, and vehicle speed measurement. They offer a very attractive alternative to inductive loops and video cameras.

### 1.1 Related Work

Traffic surveillance is an important issue for ITS. The most commonly used inductive loop detectors embedded into the road surface are considered as intrusive methods. These detectors are required to be installed underneath the road surface, which results in very high installation and maintenance costs.

Most previous works rely on the use of video cameras.[7] Using the video image signal captured by cameras, digital image processing is carried out to identify, locate, and track the target. An uncertain model adaptive model predictive control algorithm was proposed to solve the uncertainty in vehicle trajectory tracking caused by random network delay.[10] To solve the uncertainty of camera parameters, authors of[11] proposed an image-based uncalibrated trajectory tracking control scheme independent of camera parameters, without any off-line or on-line calibration of camera parameters. Vehicle occlusion and view change in the environment remain to be challenging issues in camera deployment. Using multi-camera multi-view input can solve the problems of vehicle occlusion and view change to a certain extent. Two components, multi-cross-view trajectory prediction network and cooperative correlation filter, were used in.[17] This studyrealizes the information sharing between cameras and has certain robustness to vehicle occlusion and view change.

Further author information: (Send correspondence to X. Ren)
X. Ren: E-mail: xjren@xidian.edu.cn
Y. Wang: E-mail: qq1347326386@163.com
Y. Geng: E-mail: 2104378862@qq.com

Overall, the accuracy of the camera-based approach is often affected by the environment (like fog, rain, or snow) and limited field-of-view, The setup and maintenance cost can be much higher due to the need for high-resolution cameras, storage systems for video data, lighting for nighttime operation, the cost to set up the pole on which to install the camera, etc.

Recently, magnetic sensor has attracted great attention due to its low cost and low power consumption. Magnetic sensor can measure the magnetic field change caused by moving vehicles to the geomagnetic field. Through careful data analysis of the magnetic field change, vehicle detection, vehicle classification, and vehicle speed can be estimated. Vehicle detection and classification through an improved support vector machine classifier was proposed in[4] using magnetic sensors. They used magnetic signatures to distinguish different types of vehicles, such as heavy trucks and light-wheeled vehicles. However, the algorithm is time-consuming, which requires high processing capability to get 80% to 90% accuracy. The technique proposed in[14] used four magnetic sensors for vehicle counting and vehicle speed measurement, respectively, which resulted in 95% classification accuracy. The maximum error of the speed estimates is less than 2.5% over the entire range of 18 $km/h$ to 97.2 $km/h$. In,[6] authors proposed a small magnetic sensor which is easy to install on the road. A vehicle dipole pair was used to model the moving vehicle. The disturbance of the local magnetic field was modeled by the moving vehicle, and the characteristic changes of the magnetic field waveform were analyzed to realize vehicle identification and vehicle speed estimation. Currently, acquiring vehicle speed using magnetic sensors usually requires multiple magnetic sensors.[12,15] Large scale and high-density deployment of low-cost magnetic sensors can better obtain the information and status of vehicles on the road. There are still notable challenges with the use of multiple magnetic sensors in real-world deployments. None of the existing solutions solves continuous and realtime vehicle tracking in real scenarios using ubiquitously deployed IoT sensors.

### 1.2 Contribution

In this paper, we for the first time propose a traffic surveillance system to continuously track vehicle movements in real deployments utilizing magnetic sensor measurements. In this system, multiple magnetic sensors, mounted along the roadside, are used to track various vehicles. Real-time vehicle detection data are reported from magnetic sensors, and collected into a data center via base stations. The proposed algorithms is the first to track each vehicle's trajectory in its entirety using small and ubiquitously deployed magnetic sensors, and applied to a real scenario, distinguishing itself from earlier research which use magnetic sensors for traffic surveillance, essentially vehicle counting and vehicle speed monitoring, on a small road section.

The main contributions of this article are summarized as follows.

- First of all, this paper presents a new magnetic sensor based traffic surveillance system, which is low-cost and easy to maintain.

- Second, the paper proposes a novel data association algorithm to retrieve vehicles' trajectories, and designs a related online algorithm framework.

- Finally, the paper conducts experimental evaluation and real road deployment to evaluate the performance of the proposed solution. The results show the effectiveness and efficiency of the proposed solution.

### 1.3 Paper Organization

The rest of the article is organized as follows. Section 2 introduces preliminaries including system model and vehicle motion model. Section 3 solves a vehicle trajectory tracking problem and designs both offline and online algorithms. Section 4 evaluates the algorithm performance, and Section 5 concludes the paper.

## 2. PRELIMINARIES

### 2.1 Magnetic Sensor

All vehicles have significant amounts of ferrous metals which produce a magnetic disturbance to the Earth's magnetic field. This disturbance can be detected by a magnetic sensor.[16] In[3] shows that the magnetic distribution of the magnetic flux lines when the earth's magnetic field is temporarily changed by a car. The magnetic fields are highly distorted at the wheels and slightly distorted at other parts. In this paper, we here use a low-cost magnetic sensor. The sensor module consists of a central processor and a RM3100 sensor, which is used to detect the magnetic signals. With this sensor, the magnetic signals of three axes (i.e., x, y and z) can be collected.[9] Fig. 1 shows a sample of the magnetic signal waveforms of three axes when a vehicle passes through a magnetic sensor. The breakpoints of the waveforms represent the detected vehicle entry and vehicle departure times respectively. The collected vehicle data can be pre-processed and reported via a base station, which includes the following information.

- timestamp, which indicates the time when a vehicle passing through
- magnetic field direction
- magnetic field strength
- sensor information, which includes sensor id and position information

## 2.2 System Model

As shown in Fig. 2, we consider a two-lane scenario. Given a road lane, there are a set of magnetic sensors deployed regularly at an interval between 8 - 15$m$ along the roadside, which can be labelled sequentially as

$$b_1, ... b_i, b_{i+1}, ...$$

Base stations are deployed at an interval of 1 - 1.5$km$ to transit the reported sensing data to the data center.

Given a sensor $b_i$, it collects data when a vehicle passing through, and reports to the base station at a pre-defined interval (e.g., every 30 seconds). In particular, let $X_{ik}$ represent the reported sensing data of sensor $b_i$ at time $t_k$. It contains the sensing timestamp information $t_k$, the magnetic field direction and strength information $m_k$, and the id and position information of sensor $b_i$.

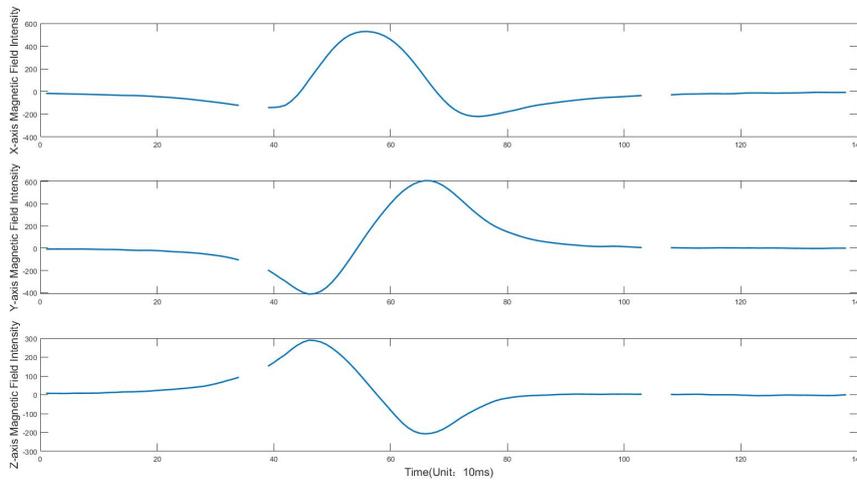

Figure 1. Sensor magnetic data

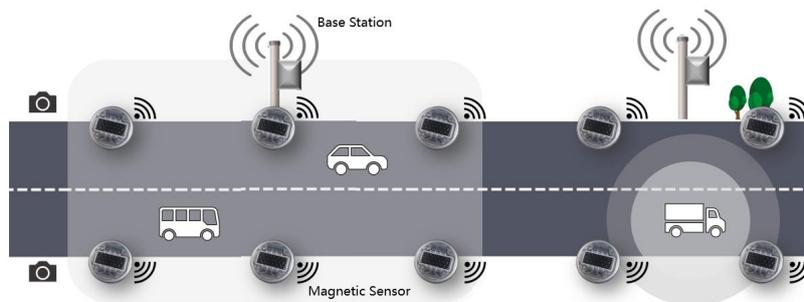

Figure 2. The system model

## 2.3 Vehicle Motion Model

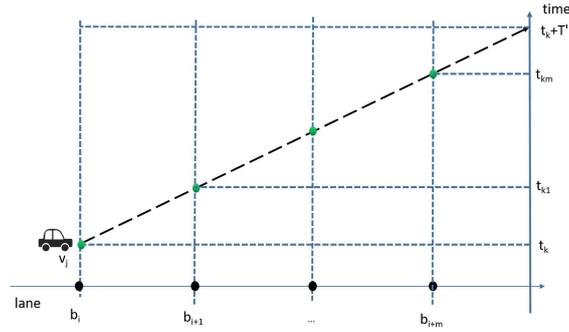

Figure 3. The vehicle model

Let $s_{jk}$ represent the velocity of vehicle $v_j$ at time $t_k$. We here consider constant-velocity (CV) motion model and constant-acceleration (CA) motion model as follows.

- CV motion: it is assumed that the vehicle travels with constant velocity in the next (typically very short) $T'$ time window.
- CA motion: it is assumed that the vehicle travels with constant acceleration in the next (typically very short) $T'$ time window. Let $a_{jk}$ represent the acceleration of vehicle $v_j$ at time $t_k$.

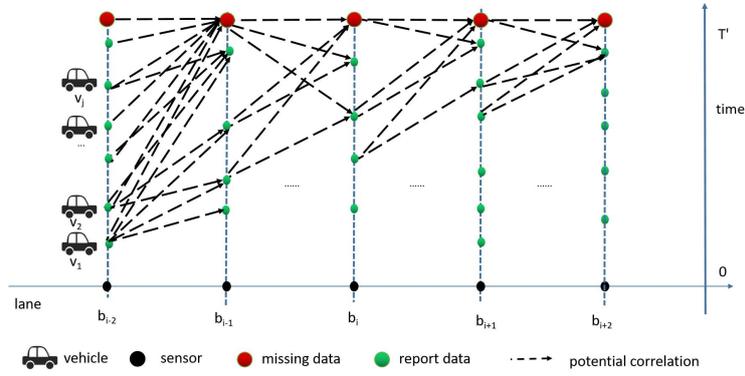

Figure 4. The time-space network

In Fig. 3, given a vehicle $v_j$ starting from sensor $b_i$ at time $t_k$ and passing through sensors $b_{i+1}, ..., b_{i+m}$ sequentially in the next $T'$ time window, we can derive that it should be sensed via following sensing data

$$\{X_{(i+1)(k1)}, ..., X_{(i+m)(km)}\}$$

Let $X_{(i+1)(k1)}$ represent the reported sensing data of sensor $b_{i+1}$ at time $t_{k1}$, and $d_{i(i+1)}$ be the physical distance between sensor $b_i$ and $b_{i+1}$. $t_{k1}$ can be calculated as follows.

- CV motion: $t_{k1} = t_k + \dfrac{d_{i(i+1)}}{s_{jk}} + \Delta t$

- CA motion: $t_{k1} = t_k + \dfrac{\sqrt{s_{jk}^2 + 2 * a_{jk} * d_{i(i+1)}} - s_{jk}}{a_{jk}} + \Delta t$

where $\Delta t$ is a random time variance among sensor devices, and is decided by the Network Time Protocol (NTP) setting.

# 3. VEHICLE TRAJECTORY TRACKING

## 3.1 Problem Definition

The vehicle trajectory tracking problem is to retrieve vehicles' trajectories based on sensing data from deployed sensors. In particular, the sensing data can be represented as follows.

$$[X_1, X_2, ... X_i, ...]$$

$$X_i = [X_{i1}, X_{i2}, ... X_{ik}, ...]$$

where $X_i$ is a set of reported data from sensor $b_i$.

The expected output is as follows.

$$[Y_1, Y_2, ... Y_j, ...]$$

$$Y_j = [Y_{j1}, Y_{j2}, ... Y_{jk}, ...]$$

where $Y_j$ represents the vehicle trajectory for vehicle $v_j$. $Y_{jk}$ represents the status of vehicle $v_j$ at time $t_k$, which includes its position info, speed info, etc.

## 3.2 Multi-commodity Network Flow Problem

Given reported sensing data $[X_1, X_2, ... X_i, ...]$ and related sensors' layout, we here define a time-space network $G = \{X' \cup B', E\}$. As per the Fig. 4, each green $b'_{ik} \in X'$ represents a sensing data from sensor $b_i$ at time $t_k$, and each red node $b'_i \in B'$ represents the related sensor $b_i$ is missing-detection. For example, there are four green nodes for sensor $b_i$, which indicates it reports four packets of sensing data. Each link in $E$ represents a potential move for a vehicle, which can be encoded using a weighted adjacency matrix $A_{|X' \cup B'| \times |X' \cup B'|}$. $a_{b'_{ik1}b'_{jk2}} \in A$ indicates a probability of a vehicle travelling from sensor $b_i$ at time $t_{k1}$ and arrives at next sensor $b_j$ at time $t_{k2}$. $a_{b'_i b'_{jk2}} \in A$ indicates a probability of a vehicle travelling from sensor $b_i$ (missing detection) and arrives at next sensor $b_j$ at time $t_{k2}$. $a_{b'_{ik1}b'_j} \in A$ indicates a probability of a vehicle travelling from sensor $b_i$ at time $t_{k1}$ and arrives at next sensor $b_j$ (missing-detection). $a_{b'_i b'_j} \in A$ indicates a probability of a vehicle travelling from sensor $b_i$ (missing detection) and arrives at next sensor $b_j$ (missing-detection). In particular,

$$a_{b'_{ik1}b'_{jk2}} = \begin{cases} 0 & \text{if } b_j \text{ not adjacent to } b_i \text{ or } k2 \leq k1 \\ f(\cdot) & \text{if } b_j \text{ adjacent to } b_i \text{ and } k2 > k1 \end{cases} \quad (1)$$

$$a_{b'_i b'_{jk2}} = p_l \quad (2)$$

$$a_{b'_{ik1}b'_j} = p_l \quad (3)$$

$$a_{b'_i b'_j} = p_l * p_l \quad (4)$$

where $d_{ij}$ is the travel distance (e.g. Euclidean distance for a straight line) between sensor $b_i$ and $b_j$. $p_l$ is an empirically estimated constant which indicates the probability of missing detection. $f$ is a convex function which indicates the probability of related data association based on vehicle velocity/acceleration info and $d_{ij}$.

Given the above time-space graph, we can formulate a multi-commodity network flow problem,[13] where passing vehicles are the set of commodities with demand of 1. Each sensor data node has capacity of 1. Each sensor 'missing-detection' node has capacity of $\infty$, for which each inbound link represents a potential vehicle missing detection. The objective is to maximize the flow weight, i.e., maximizing the joint possibility of data association.

### 3.3 Data Association Algorithm

If we only consider the next one hop for each vehicle, the formulated multi-commodity network flow problem can be reduced to a maximum weight bipartite matching problem.[1] We devise a data association algorithm accordingly to jointly detect and track vehicles.

---

**Algorithm 1:** G_Association

---

**Input**  : reported sensing data
           given time-window
           sensor deployment information

**Output** : vehicles' trajectories
Pre-process the sensing data including removing false-sensing data;
Sort the sensing data by the sensor index;
**for** <u>each unprocessed sensor</u> **do**
    /* track vehicles                                                           */
    **for** <u>each existing vehicle</u> **do**
        Get the latest vehicle status;
        Calculate the related weight for each data-link;
        /* for simplicity, we only consider one-hop here                       */
    **end**
    Solve the formulated multi-commodity problem to match sensor data to vehicles;
    /* Since we only consider one-hop, we can solve a maximum weight bipartite matching
    problem via KM algorithm[8]                                                 */
    Apply Kalman filter (CV or CA motion model) to update vehicle status;
    **for** <u>each un-associated sensing data for this sensor</u> **do**
        /* detect vehicles                                                     */
        Initialize a new vehicle;
    **end**
**end**

---

THEOREM 3.1. *Given reported sensing data* $[X_1, X_2, ..., X_i, ...]$, *there is a data association algorithm G_Association for the vehicle trajectory tracking problem, which takes* $O(Max(|X_1|, |X_2|, ..., |X_i|, ...)^3)$ *time.*

*Proof.* We analyze the time complexity of algorithm *G_Association* as follows. With each iteration $i$ except the first iteration, it takes $O((|X_{i-1}| + |X_i|)^3)$ time to detect and track vehicles. For the first iteration, it takes $O(|X_1|)$ time, since it only needs to initialize vehicles. The number of iterations is determined by the number of sensors, which is a constant. Hence the algorithm takes $O(Max(|X_1|, |X_2|, ..., |X_i|, ...)^3)$ time.

### 3.4 Algorithm Framework

Given that sensing data are reported in a real-time manner for real-world scenario, vehicle trajectories need to be updated continuously. To this end, we here propose an online algorithm framework to generate vehicle trajectory data stream (e.g. every 30 seconds). Overall, the proposed algorithm framework caches the reported data, calls the proposed data association algorithm *G_Association* to form updated vehicle trajectories, and pushes the updated result if required (e.g. for visualization) every pre-configured time interval.

| Algorithm 2: G_Algorithm_Framework |
|---|
| **Date:** reported sensing data |
| **Result:** vehicle trajectories |
| Initial empty vehicle store; |
| Initial empty data store to save the reported sensing data; |
| /* real-time data is saved into data store via rest-api interface       */ |
| Initial a pre-configured timer (e.g. 30 seconds); |
| **while** <u>timer trigger</u> **do** |
|     Fetch reported sensing data from data store; |
|     Fetch un-completed vehicles from vehicle store; |
|     Call **G_Association** for vehicle detect and track; |
|     Update vehicle store; |
|     Push the updated data if required; |
| **end** |

## 4. PERFORMANCE EVALUATION

In this section we study the performance of the proposed solution through both experimental simulation and real-world road deployment.

### 4.1 Experimental Simulation

A two-lane experiment was conducted on Zhangba East Road in Xi'an city. As shown in Fig. 5(a), the width of a single lane is 3.5$m$. Magnetic sensors are evenly deployed along the lane lines on both sides, and the deployment interval of magnetic sensors on the same lane is 15$m$. There are many types of vehicles on both lanes, including cars, buses, trucks, sprinklers, single-layer buses, double-layer buses, etc. The magnetic field waveforms of these vehicles passing through magnetic sensors are different, where the magnetic field waveforms of three axes (i.e., x, y and z) changes when a vehicle passes through a magnetic sensor.

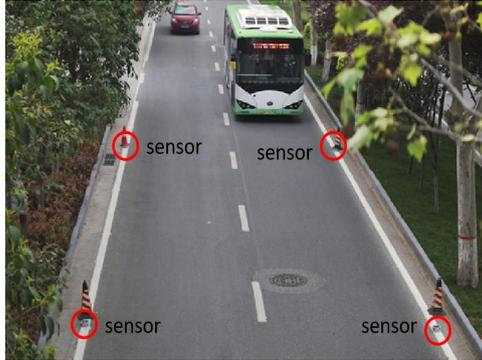
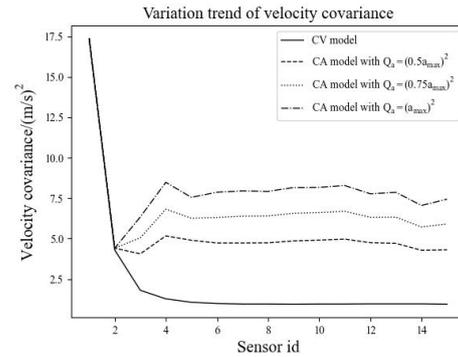

(a) Simulation environment          (b) Comparison of CV and CA motion models

Figure 5. The simulation

    The detected vehicle data are pre-processed by the microprocessor connected to the magnetic sensors to obtain the timestamp, magnetic field direction, magnetic field strength and other related information. These information are collected via laptop, and used as experiment data. At the same time, a camera is deployed together. The camera data within the same time window is collected, and processed manually to extract vehicle trajectories, which are used as 'ground of truth'.

    We first evaluate two different vehicle motion models by comparing the related Kalman filter velocity covariance in single-vehicle scene. The derivative of acceleration adopted in CA motion model is a white Gaussian noise with mean value of zero, and its variance $Q_a$ is a constant which does not change with time. Generally, $\sqrt{Q_a}$ can be empirically

selected as $0.5a_{max} - a_{max}$ without considering other errors, where $a_{max}$ is the fastest acceleration of an ordinary vehicle from 0 to 100 kilometers per hour. It can be seen from Fig. 5(b) that the velocity covariance of the CV motion model finally converges to around 1. When the $Q_a$ value is $(a_{max})^2$, the velocity covariance of the CA motion model converges to around 7.5, which is much worse than the CV motion model. With the decrease of $Q_a$ value, the variance of velocity also decreases. When the $Q_a$ value is $(0.5a_{max})^2$, the velocity covariance of the CA model converges to about 4.5, which is still worse than the CV motion model. Overall, the velocity variance of the CA motion model is much worse than that of the CV motion model. It should be due to that the data we can measure is the position of the vehicle and the related timestamp, but the speed information of the vehicle is missing. To this end, we adopt the CV motion model to continue the rest of experiment.

We then evaluate the track accuracy by comparing the camera vehicle trajectory data with the generated vehicle trajectory data. The experimental result shows that the track accuracy of the algorithm *G_Association* is larger than 95%.

## 4.2 Real Road Deployment

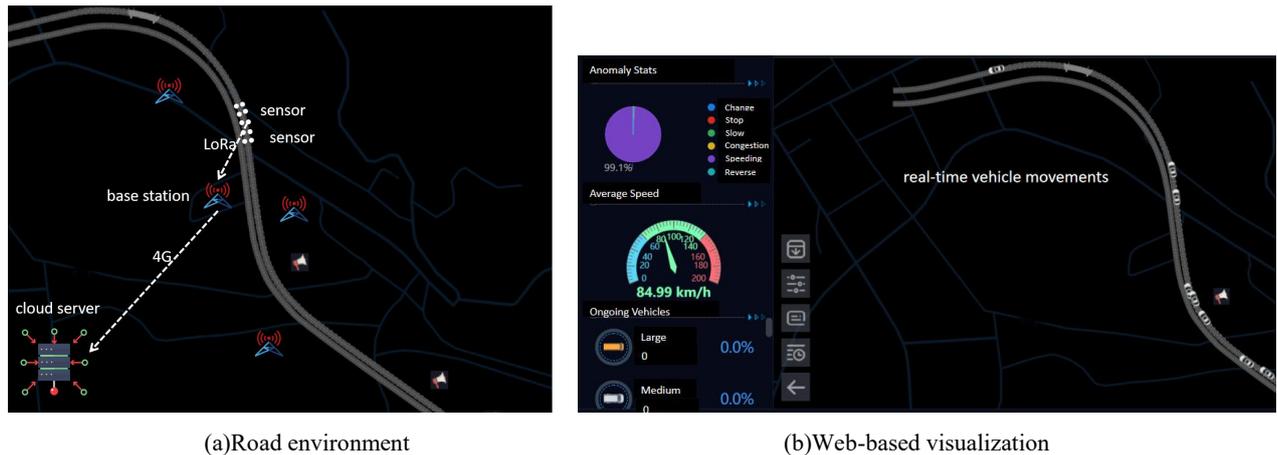

(a)Road environment                                     (b)Web-based visualization

Figure 6. The real deployment

A real-world road deployment was conducted on Qingyuan-Lianzhou Highway. As shown in Fig. 6(a), around 300 magnetic sensors are deployed along the $1.64km$ lane lines on both sides, and the average deployment interval of magnetic sensors on the same lane is around $15m$. There are 4 base stations deployed along the road to collect the reported data. As per the motion model comparison result, we here adopt the CV motion model accordingly.

A web-based portal is deployed for visualization, where vehicle trajectories are pushed to the portal every 30 seconds as shown in Fig. 6(b). Similar as the simulation, we use the camera data as 'ground truth' to validate the solution performance, which delivers more than 95% accuracy. In Fig. 6(b), together with the real-time vehicle trajectories tracking, the system is capable of addressing vehicle counting, vehicle speed and traffic anomaly detection (e.g. speeding, parking, etc.).

## 5. CONCLUSION AND FUTURE WORK

Traffic surveillance is of vital importance for Intelligent Transportation Systems. This study contributes to investigate traffic surveillance via low-cost magnetic sensors, model and track vehicle trajectories of road networks using a graph-based data association algorithm, and design a related online algorithm framework. The proposed solution in this paper has efficient performance in both simulation and real deployment, which can benefit to vehicle surveillance via a more efficient approach in terms of cost, reliability, etc. In the future, much complicated scenarios as follows will be investigated.

- to investigate on multi-lane roads
- to investigate on crossroads

## ACKNOWLEDGMENTS

This work was supported by the National Natural Science Foundation of China under Grant U21A20446; Guangzhou Applied Basic Research Program under Grant 202201011786.